\documentclass[11pt]{article}

\usepackage[margin=1in]{geometry}
\usepackage{amsmath}
\usepackage{booktabs}
\usepackage{graphicx}
\usepackage{xcolor}
\usepackage[most]{tcolorbox}
\tcbuselibrary{listings}
\usepackage[colorlinks=true,linkcolor=blue,citecolor=blue,urlcolor=blue]{hyperref}
\usepackage{enumitem}
\usepackage{fontspec}
\usepackage{xeCJK}
\usepackage{microtype}
\usepackage[round,authoryear]{natbib}
\newtcblisting{pythonbox}{
    listing only,
    listing options={language=Python, basicstyle=\ttfamily\small,
        keywordstyle=\color{blue!70!black}, commentstyle=\color{gray},
        stringstyle=\color{teal}, breaklines=true, showstringspaces=false},
    colback=gray!5, colframe=gray!40, boxrule=0.4pt, arc=2pt,
    left=6pt, right=6pt, top=4pt, bottom=4pt,
}
\newcommand{\model}[1]{\textsc{#1}}
\newcommand{\dataset}[1]{\texttt{#1}}
\newcommand{\promptlen}{\text{prompt\_len}}

\title{Nanbeige4.2-3B on Apple Silicon: Fixing Deployment Bugs and Decreasing Looped Transformer Memory Overhead}

\author{John T.\ Halloran \\ \texttt{halloj3@uw.edu}}
\date{\today}

\begin{document}

\maketitle

\begin{abstract}
\href{https://huggingface.co/Nanbeige/Nanbeige4.2-3B}{\model{Nanbeige4.2-3B}}~\citep{nanbeige2026} is a 3B-parameter agentic model built around a
Looped Transformer (LT)~\citep{bae2026mor} that reuses one stack of layers for a second
forward pass, adding effective depth without additional parameters.  Evaluated on Apple Silicon (MPS), we identify
five independent bugs which prevent the released checkpoint from running via Hugging Face \texttt{transformers}
out of the box (including a silently-zeroed RoPE buffer and calls to removed \texttt{transformers} cache APIs).
Furthermore, we show that fixing these bugs is still not sufficient for agentic tasks, due
to the LT's layer-reuse strategy (which effectively doubles peak attention memory) used to achieve parameter
efficiency.  We thus introduce a chunked-prefill strategy which alleviates the incurred memory-capacity penalty,
extending allowable context width by $2.7 \times$ on 32~GiB shared memory.  However, even with the reduced memory
overhead, we show that patches are required to render \model{Nanbeige4.2-3B} usable; resolving both system prompt and MPS-native memory bugs finally allows reliable evaluation on standard MCP and tool-calling benchmarks.
On a subset of MCPMark, the debugged model completes up to \textbf{30\%} of real agentic tasks
(up from the original's 0\%), while, on BFCL, it is near-perfect at single tool calls
(yet fails the majority of multi-tool tests).
We release the patched checkpoint, system prompt optimizer, and evaluation harnesses at \href{https://github.com/johnhalloran321/Nanbeige4.2-3B-mps-fix}{\texttt{github.com/johnhalloran/\allowbreak Nanbeige4.2-3B-mps-fix}}.
\end{abstract}

\section{Introduction}

\model{Nanbeige4.2-3B} has recently been released as a capable small language
model (SLM) specifically designed for improved capability at agentic tasks. The
model utilizes a Loop Transformer (LT) architecture to trade off decoding
compute without requiring scaling parameter count. The released model card
reports competitive or better results than larger models (Qwen3.5-4B,
Qwen3.5-9B) on agentic and office-workflow benchmarks~\citep{nanbeige2026}, credited to the added
effective depth of its Looped Transformer architecture. However, running the
released checkpoint in a ReAct-style agentic harness via \texttt{transformers}
on MPS (Apple Silicon) surfaces both stability and correctness problems rendering the model unusable out of the box.

Herein, we identify five initial bugs---e.g., a silently-zeroed RoPE buffer, calls to removed
\texttt{transformers} cache APIs, etc.---and the necessary fixes to render the released checkpoint usable on MPS.
However, the resulting model remains unscalable for agentic tasks due to the increased memory needs of the LT architecture, despite
the 3B parameters in \texttt{bf16} on a 32~GiB shared memory system; inherently, the LT enables parameter efficiency
at the cost of double the peak attention memory (due to the recursive looping over layers in the forward pass), which prohibits the
long reasoning traces required for agentic tasks.  Thus, we introduce a chunked-prefill strategy which relieves LT memory overhead,
more than doubling allowable context-width evaluation.
However, the resulting model reveals further deficiencies: (1) \model{Nanbeige4.2-3B}'s trained-in tool-use system
prompt is silently replaced, not merged, the moment a caller supplies any system message, and (2) an MPS out-of-memory (OOM) error can permanently degrade the serving process's
usable memory budget, surfacing only while evaluating the debugged model on MCPMark.  The gamut of bug and model fixes---five initial bug fixes, an alternative prefilling algorithm, system prompt correction, and MPS-specific OOM
fix---allow the true reproducible evaluation of the model on standard agentic and tool-use benchmarks.

\section{Five Initial Deployment Bugs}~\label{sec:bugs}

Loading on Apple Silicon via
\begin{pythonbox}
from transformers import AutoModelForCausalLM
model = AutoModelForCausalLM.from_pretrained("Nanbeige/Nanbeige4.2-3B",
                                             trust_remote_code=True,
                                             device="mps",
                                            )
\end{pythonbox}
fails or silently misbehaves for five independent
reasons. We confirm each by direct reproduction against the unmodified checkpoint.

\begin{enumerate}[label=\arabic*.,leftmargin=*]
\item \textbf{RoPE buffer persistence (dominant bug).} The model's \texttt{inv\_freq}
  rotary-embedding buffer is silently zeroed on load and never repopulated before
  the first forward pass. RoPE therefore contributes zero positional
  information to attention: \textbf{the model runs without knowing token order}. This
  manifests as fluent-looking but positionally-incoherent generation rather than
  a crash, which makes it easy to miss without directly inspecting buffer values
  after load.

\item \textbf{RoPE-config dispatch \texttt{KeyError}.} A bug
  in the custom modeling code's RoPE-type dispatch raises \texttt{KeyError} for a
  subset of otherwise valid config values (Table~\ref{tab:bugs}). This fires during model construction, before
  device placement or a single forward pass, and is not MPS-specific (it
  blocks loading on any device).

\item \textbf{Cache API sentinel mismatch.} Calling the model's \texttt{forward()} directly
  with the default \texttt{past\_key\_values=None}, instead of through \texttt{generate()},
  calls an API removed in current \texttt{transformers} releases (\texttt{DynamicCache.from\_legacy\_cache(...)}).

\item \textbf{Position-IDs re-trim.} A bug during position-tracking in the custom attention
  code produces a hard crash on device MPS specifically (does not reproduce on CPU).

\item \textbf{Tied-weights key format.} An incompatible tied-weights key naming
  convention breaks \texttt{save\_pretrained()}.  Even a successfully-patched,
  running model cannot be re-serialized without an additional fix.
\end{enumerate}

We fix all five via sibling-file monkeypatching (never modifying cached
\texttt{transformers} package files) in the
\href{https://huggingface.co/johnhalloran/Nanbeige4.2-3B-mps-fix}{\texttt{johnhalloran/Nanbeige4.2-3B-mps-fix}} checkpoint. None of the five is
related to the memory or system-prompt issues in
Sections~\ref{sec:tradeoff}--\ref{sec:sysprompt}, which persist after all five
are fixed.

Table~\ref{tab:bugs} gives the exact trigger, line number in the unmodified
\texttt{modeling\_nanbeige.py}, and error string or symptom for each bug,
against \texttt{transformers==5.8.1} (the version used throughout this paper);
full diffs and reproduction scripts are in the artifact repository's
\texttt{patch/} directory (Section~\ref{sec:artifact}).

\begin{table}[h]
  \centering
\caption{Exact trigger, source line, and error/symptom for each bug, against
\texttt{transformers==5.8.1}. Line numbers refer to the unmodified checkpoint.}
\label{tab:bugs}
\footnotesize
\begin{tabular}{@{}p{0.16\linewidth}p{0.03\linewidth}p{0.32\linewidth}p{0.4\linewidth}@{}}
\toprule
Bug & Line & Trigger & Error / symptom \\
\midrule
1. RoPE buffer (major) & 947 & \texttt{from\_pretrained()} + first \texttt{forward()} & no exception raised; \texttt{inv\_freq} buffer reads all zeros (\texttt{persistent=False}, never restored after a meta-device load) \\
2. RoPE-config dispatch & 1077 & \texttt{from\_pretrained(..., trust\_remote\_code=True)} & \texttt{KeyError: 'type'} \\
3. Cache API sentinel & 2125 & calling \texttt{forward()} directly with \texttt{past\_key\_values=None}, bypassing \texttt{generate()} & \texttt{AttributeError: type object 'DynamicCache' has no attribute 'from\_legacy\_cache'}; still present in the deployed checkpoint \\
4. Position-IDs re-trim & 2630 & \texttt{generate()} on MPS, once \texttt{prepare\_inputs\_for\_generation} receives an already-populated \texttt{position\_ids} kwarg & \texttt{'mps.matmul' op contracting dimensions differ 361 \& 181} (Metal assertion, uncatchable \texttt{SIGABRT}); not reproduced on CPU \\
5. Tied-weights key format & 2417 & \texttt{model.save\_pretrained(...)} & \texttt{AttributeError: 'list' object has no attribute 'keys'} \\
\bottomrule
\end{tabular}
\end{table}

\section{The Looped-Transformer Memory Tradeoff}
\label{sec:tradeoff}

\model{Nanbeige4.2-3B}'s LT~\citep{bae2026mor} feeds the hidden state
through the full stack of $L$ physical transformer layers, then re-feeds the
output of that pass through the same $L$ layers a second time before producing
logits: two effective passes ($2L$ effective layer-executions) from $L$ layers'
worth of parameters.  This effectively improves model quality without scaling parameters---e.g., LTs
outperforming larger non-looped models at a fixed parameter budget~\citep{bae2026mor}.  However,
parameter counts are kept low at the expense of additional compute and memory requirements.

Naively, self-attention's peak activation memory during prefilling is dominated by
materializing an attention-score tensor proportional to $(\text{prompt\_len})^2$ per layer
pass.  Compared to their non-looped counterparts, LTs double required prefilling memory, as 
the $O(\text{prompt\_len}^2)$ attention computation over
the same prompt is repeated once per loop.  Thus, despite the total-parameter count savings
enabled by looped weight-sharing during pretraining, naive prefilling results in
twice the full quadratic attention cost during inference.

For small language models (SLMs) on large dedicated hardware (e.g., H200s), this doubling
is usually not an issue.  However, on Apple Silicon's unified memory---shared between the OS,
competing processes, and the model with no page-out path comparable to CUDA's memory
management---it can be catastrophic for agentic tasks.

\subsection{Balancing Looped Memory Use via Chunked Prefilling}\label{sec:chunked}
As opposed to \emph{naive prefilling}---wherein the full
$(\text{prompt\_len} \times \text{prompt\_len})$ attention-score tensor is computed
once per loop iteration in a single forward pass---we show that \model{Nanbeige4.2-3B}'s
memory use may be significantly decreased via \emph{chunked prefilling}.  
Chunked prefilling processes the prompt in fixed-size chunks, growing the Key-Value cache
incrementally between chunks the same way ordinary autoregressive decoding already does.

We replace the single-shot \texttt{model(input\_ids=full\_prompt, ...)} prefill call with a loop
that processes the prompt in fixed-size chunks (256 tokens by default), incrementally growing a \texttt{DynamicCache}
between chunks, before handing the remainder off to \texttt{generate()}:
\begin{pythonbox}
def _chunked_prefill_generate(model, input_ids, max_new_tokens, chunk_size=256, **gen_kwargs):
    total_len = input_ids.shape[1]
    if total_len <= chunk_size:
        return model.generate(input_ids=input_ids, max_new_tokens=max_new_tokens, **gen_kwargs)
    cache = DynamicCache()
    n_full_chunks = (total_len - 1) // chunk_size
    with torch.no_grad():
        for i in range(n_full_chunks):
            start, end = i * chunk_size, i * chunk_size + chunk_size
            outputs = model(
                input_ids=input_ids[:, start:end], past_key_values=cache,
                use_cache=True, cache_position=torch.arange(start, end),
            )
            cache = outputs.past_key_values
    return model.generate(input_ids=input_ids, past_key_values=cache,
                           max_new_tokens=max_new_tokens, **gen_kwargs)
\end{pythonbox}

This bounds the peak per-step attention-score tensor to
$(\text{chunk\_size} \times \text{running-total})$, independent of how long the prompt is,
at the cost of splitting one forward pass into several sequential sub-calls instead of one.
Bit-identical outputs were verified against naive prefill.

\subsection{LongBench-Pro results}
We demonstrate the memory benefits of chunked prefilling (CP) over naive prefilling (NP)
using 50 long samples from \dataset{LongBench-Pro}~\citep{longbenchpro} and single-turn queries of
eight lengths varying from 1024 to $12,244$ tokens.  Each sample is tokenized
using the \model{Nanbeige4.2-3B} tokenizer and truncated to the target length.  To measure
maximum memory throughput, we calculate the max batch size per length and prefilling strategy by doubling
the batch size until failure, repeating this process 3 times.  All evaluations were performed on a
\texttt{Apple M2 Max} with 32~GiB of shared memory.
Results are in Table~\ref{tab:memory}.

\begin{table}[h]
\centering
\caption{NP vs CP evaluated over 50 LongBench-Pro samples, averaged over 3 repeated experiments.
``--'' denotes the method could not complete even batch=1.
}
\label{tab:memory}
\begin{tabular}{rrrrr}
\toprule
$\promptlen{}$ & NP max batch & NP tok/s & CP max batch & CP tok/s \\
\midrule
1024      & 16 & $275.3$ & 32 & $212.5$ \\
2048      & 4  & $267.5$ & 16 & $158.1$ \\
4096      & 2  & $158.4$ & 4  & $124.8$ \\
8192      & -- & --      & 1  & $168.7$ \\
9205      & -- & -- & 1 & $156.5$ \\
10218     & -- & --      & 1  & $148.6$ \\
11231     & -- & --      & 1  & $142.4$ \\
12244     & -- & --      & -- & -- \\
\bottomrule
\end{tabular}
\end{table}

The maximum length possible under CP (11231) is significantly larger than NP (4096).
However, CP trades memory requirements for time; at $\promptlen{}=1024$, CP allows
twice the amount of batch-parallelism, while only being $22.8\%$ slower than NP.
This tradeoff is most noticeable at $\promptlen{}=2048$, where CP allows 4 times the
batch-parallelism while being $40.9\%$ slower.  We note that, for CP, per-chunk subcalls
incur a fixed cost, paid $\text{chunk\_size}$ times regardless of batch size.  Thus, this overhead
becomes amortized when the batch size is large, but becomes a larger portion of total runtime when
only smaller batch sizes are present, e.g., $\promptlen{}=4096$, which achieves lower throughput
than batch size $= 1$ evaluations over longer prompts.  We note that folding the sequential
per-chunk subcalls into fewer large GPU operations via kernel fusion would reduce this per-call
overhead directly.

\section{System-Prompt Regression}~\label{sec:sysprompt}

Independent of the previously discussed memory issues, \model{Nanbeige4.2-3B}'s chat template
(\texttt{chat\_template.jinja}, in the \texttt{\{\% if tools \%\}} branch) performs the following
if/else on \texttt{messages[0]}:

\begin{itemize}[leftmargin=*]
\item If the caller supplies \textbf{any} system message, it is used verbatim, with a
  trailing \texttt{"\textbackslash n\textbackslash n"} the template appends.
\item Otherwise, the template injects a hardcoded default (Nanbeige's own trained-in
  tool-use system prompt, beginning ``你是一位工具函数调用专家...''\footnote{Translation:
  ``You are a tool-function-calling
  expert. You will be given a question and a set of possible tool functions. Based
  on the question, you need to make one or more function/tool calls to accomplish
  the goal --- please do your best to explore solving the problem through tools. If
  no function is usable, reply directly to the user in natural language. If the
  given question is missing parameters required by a function, use natural language
  to ask the user for the necessary information. If the call results are already
  sufficient to answer the user's question, summarize the results and reply to the
  user in natural language.''}) with \textbf{no} trailing separator before the
  \texttt{\# Tools} section that follows.
\end{itemize}

Any caller-supplied system message
thus silently discards the model's own trained default instead of extending it.  Any
tools-plus-user-message request produces a clean, correctly-formatted single tool call, but with
no system message; if a system message is added---even one such as ``You are an assistant with
MCP tools''---multi-tool-call outputs break into malformed text.

Re-supplying the original text as an explicit system message
does not fix this behavior. Byte-identical content through the explicit-system-message
branch still breaks, because that branch's own auto-appended
\texttt{"\textbackslash n\textbackslash n"} differs from the zero-extra-whitespace
auto-insert branch's output by exactly two characters. The released checkpoint's
tool-calling reliability is calibrated to the exact byte sequence its own
default rendering path produces, consistent with its tool-use SFT/RL data having
only ever been rendered through the auto-insert branch and never with a
caller-supplied system message.

\textbf{Fix:} remove the system message from the chat template, such that the template
takes its own zero-extra-whitespace auto-insert path.  Then insert the caller's
system content in the rendered string after the auto-insert default (never through
the default template's aforementioned branch).  This generates single-tool-call
outputs while including the caller's system content.

We note that a closely related (but a mechanically distinct bug) has been independently reported
against this model:
\href{https://github.com/ggml-org/llama.cpp/pull/26324}{llama.cpp PR \#26324}
documents Nanbeige4.2-3B emitting \texttt{<tool\_call>} with a trailing space
instead of \texttt{<tool\_call>\textbackslash n} for roughly 25\% of calls,
breaking tag-matching in that inference engine, and notes the same template
structure is shared by (though not observed to trigger the same failure in)
Qwen3-Coder and Qwen3.5-4B. Both bugs sit in the same chat-template/generation
pipeline and are independent evidence that this checkpoint's tool-calling
reliability is template and whitespace sensitive.

\section{Evaluation}
\label{sec:eval}

We evaluate the combined fixes on a 10 task subset of \dataset{MCPMark}~\citep{mcpmark}---which tests
MCP-tool use capabilities on multi-turn tasks while grading tool-generated responses---and a 150 subset of the \dataset{Berkeley Function-Calling Leaderboard} (\dataset{BFCL})~\citep{bfcl}--which tests
tool selection ability without considering tool execution outputs.

\textbf{MPS memory bug.} Running the test suite end-to-end surfaced a bug unrelated to the model: a single caught \texttt{RuntimeError: MPS backend out
of memory} permanently degrades the harness process's usable MPS memory
budget for the rest of its life. Neither \texttt{torch.mps.empty\_cache()}
nor \texttt{gc.collect()} reclaim it in a controlled tests; only a process restart does. Uncorrected,
one task's OOM---expected for \dataset{MCPMark}, where multi-turn tasks can grow past
the max-token ceiling ($12,244$, per Table~\ref{tab:memory})---cascades into spurious OOMs on every later, unrelated task in the
same long-lived server. Restarting the harness fresh before each task
(mirroring the subprocess-per-trial isolation already used for
Table~\ref{tab:memory}) fixes this.

\subsection{\dataset{MCPMark} (Filesystem subset, easy tier)}
\label{sec:mcpmark-diagnosis}

\begin{table}[h]
\centering
\caption{\dataset{MCPMark} Filesystem (easy tier), patched checkpoint, per-task server isolation, using
  \dataset{MCPMark}'s default 1 hour timeout.}
\label{tab:mcpmark}
\begin{tabular}{lrl}
\toprule
Task & Turns & Outcome \\
\midrule
largest\_rename     & 5 & \textbf{pass} \\
txt\_merging        & 5 & \textbf{pass} \\
file\_reorganize    & 7 & \textbf{pass} \\
pattern\_matching   & 2 & fail, timed out \\
file\_splitting     & 4 & fail, tool response OOM \\
uppercase           & 7 & fail, tool response OOM \\
structure\_analysis & 2 & fail, tool response OOM \\
papers\_counting    & 2 & fail, tool response OOM \\
duplicate\_name     & 2 & fail, tool response OOM \\
recommender\_name   & 2 & fail, tool response OOM \\
\bottomrule
\end{tabular}
\end{table}

Evaluations were run over \dataset{MCPMark}'s Filesystem suite of 10 easy tasks (which do not require
API credentials), using
the benchmark's default timeout of 1 hour.  The original (unpatched) checkpoint cannot be evaluated
(on any device) due to bug 2 of Section~\ref{sec:bugs}.
The patched checkpoint---with CP (Section~\ref{sec:chunked}) and all described
bug fixes (Sections~\ref{sec:bugs} and ~\ref{sec:sysprompt})---scores \textbf{3/10 (30\%)}.  Details
for each per-task run are in Table~\ref{tab:mcpmark}.

For the failed pattern\_matching task, the model correctly calls the MCP tool \texttt{read\_multiple\_files} once, but repeats a long absolute path 21 times, leading to a long context-width that eventually times out.  The remaining failures accumulate context over multiple turns and eventually exceed memory capacity before solving the underlying task.
\subsection{Tool-calling correctness in isolation from decode throughput (\dataset{BFCL})}
\label{sec:bfcl}
To stress test the tool calling ability of the model (and not necessarily end-to-end task effectiveness), 
we evaluate the debugged/scalable model on a 150 task subset of \dataset{BFCL}'s non-live, single-turn categories
\citep{bfcl}: \texttt{simple\_python} (one correct call), \texttt{multiple} (pick 1 of $N$ candidate functions),
\texttt{parallel} (emit 2+ calls to the same function), \texttt{parallel\_multiple}
(2+ calls to different functions), and \texttt{irrelevance} (correctly emit no
call at all). These are AST/exact-match graded, require no external LLM judge,
and complete in one generation each (10--20s), so there is no wall-clock or
multi-turn-accumulation confound (as in \dataset{MCPMark}).

\begin{table}[h]
\centering
\caption{\dataset{BFCL}, patched checkpoint, 30 tasks per category.}
\label{tab:bfcl}
\begin{tabular}{lrl}
\toprule
Category & Score & Dominant failure mode \\
\midrule
\texttt{simple\_python}     & 19/30 (63.3\%) & wrong call count (11/30) \\
\texttt{multiple}           & 13/30 (43.3\%) & wrong call count (17/30) \\
\texttt{irrelevance}        & 30/30 (100\%)  & --- \\
\texttt{parallel}           & 1/30 (3.3\%)   & wrong \# of functions (29/30) \\
\texttt{parallel\_multiple} & 9/30 (30.0\%)  & wrong \# of functions (21/30) \\
\bottomrule
\end{tabular}
\end{table}

The model reliably recognizes when \emph{not} to call a tool (100\% on
\texttt{irrelevance}) and is moderately reliable on a single, well-specified
call (63.3\%). It is specifically weak at emitting
\emph{multiple} tool calls in one turn: on both parallel categories, the
dominant failure is producing the wrong number of function calls, almost
always one call where two were required. This is a distinct, format-level
limitation from anything in Section~\ref{sec:mcpmark-diagnosis}---it would
appear even on hardware with sufficient memory capacity to avoid OOMs and
speed to avoid timeouts.

\section{Conclusions and Artifacts}
\label{sec:artifact}

The released \model{Nanbeige4.2-3B} checkpoint cannot run reliably via Hugging Face
\texttt{transformers} on Apple Silicon, blocked first by five independent deployment
bugs and then by the memory overhead of its Looped Transformer architecture. We fix
all five bugs, introduce chunked prefilling to more than double the usable context
width under Apple Silicon's shared-memory limits, and resolve a system-prompt
regression and an MPS memory bug that otherwise block reliable agentic evaluation.
The resulting patched checkpoint completes up to 30\% of a real MCPMark agentic
subset (up from 0\%) and is near-perfect at single BFCL tool calls, though it still
fails the majority of multi-tool-call tests.

The code, patched checkpoint, and reproduction scripts for all evaluations in are released at \href{https://github.com/johnhalloran321/Nanbeige4.2-3B-mps-fix}{\texttt{github.com/johnhalloran321/\allowbreak nanbeige-mps-fix}}
(repository pending publication; layout and contents already finalized, see
its \texttt{README.md}) and at
\href{https://huggingface.co/johnhalloran/Nanbeige4.2-3B-mps-fix}{\texttt{johnhalloran/\allowbreak Nanbeige4.2-3B-mps-fix}}
on Hugging Face.

\bibliographystyle{plainnat}
\bibliography{references}

\end{document}